\ifdefined\XeTeXversion\else \pdfoutput=1 \fi
\documentclass[10pt,twocolumn,letterpaper]{article}
\usepackage[pagenumbers]{cvpr} %

\usepackage{cuted} %

\usepackage{currfile} %

\usepackage{caption} %

\definecolor{roleHarness}{RGB}{57,82,112}
\definecolor{roleAgent}{RGB}{111,78,103}
\definecolor{roleModel}{RGB}{51,103,91}
\definecolor{roleTool}{RGB}{126,94,57}
\newcommand{\tableHarness}[1]{\textcolor{roleHarness}{#1}}
\newcommand{\tableAgent}[1]{\textcolor{roleAgent}{#1}}
\newcommand{\tableModel}[1]{\textcolor{roleModel}{#1}}
\newcommand{\tableTool}[1]{\textcolor{roleTool}{#1}}

\usepackage{multirow}
\usepackage{listings}
\usepackage{algorithm}
\usepackage{algorithmic}
\graphicspath{{figures/}}
\definecolor{cvprblue}{rgb}{0.21,0.49,0.74}
\usepackage[pagebackref,breaklinks,colorlinks,allcolors=cvprblue]{hyperref}
\def\confName{CVPR}
\def\confYear{2027}

\newcommand{\sys}{\textsc{SceneAnything}}

\title{\sys: Agentic Reconstruction of Large-Scale Mesh Scenes from a Single Video}

\author{Bowen Xue$^{1}$ \qquad Saeed Hadadan$^{1}$ \qquad Fabrice Rousselle$^{1}$\\[2pt] Zahra Montazeri$^{2}$ \qquad Ravi Ramamoorthi$^{3,1}$ \qquad Milos Hasan$^{1}$\\[6pt] $^{1}$NVIDIA \qquad $^{2}$University of Manchester \qquad $^{3}$University of California, San Diego}

\newcommand{\supppage}[1]{\clearpage\thispagestyle{empty}%
  \vspace*{-\dimexpr 1in+\topmargin+\headheight+\headsep+\topskip\relax}\nointerlineskip
  \noindent\hspace*{-\dimexpr 1in+\oddsidemargin\relax}%
  \raisebox{-\height}[0pt][0pt]{\includegraphics[page=#1,width=\paperwidth]{supp.pdf}}\par}
\begin{document}

\maketitle
\begin{strip}
\centering
\includegraphics[width=\textwidth]{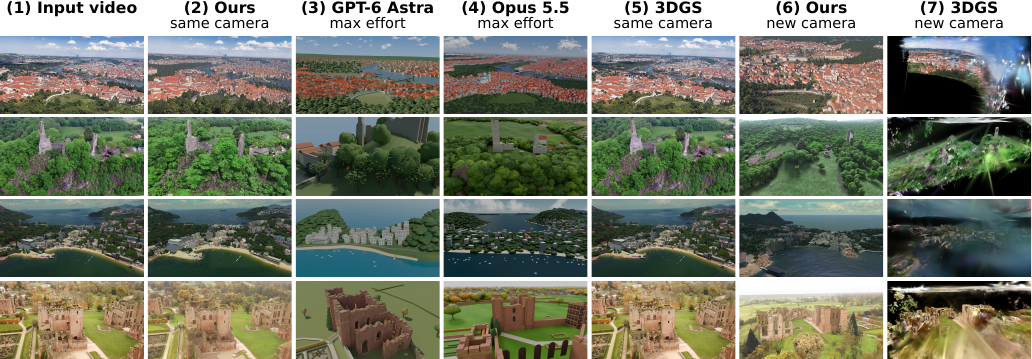}
\captionof{figure}{\sys{} turns a casually recorded or generated video into a 3D scene made of separate, editable mesh objects. Each row is one video. The agents of columns~3 and~4 are each prompted directly to reconstruct the video using Blender; 3DGS~\cite{kerbl2023gaussians} is trained on the video without the frame of column~1. Working alone, the agents produce plausible objects, but with simplified shapes and suboptimal layout, scale and lighting. Our method achieves a much better fit in terms of geometry, materials and lighting. As expected, 3DGS fits the video well (column~5) but fails at a new camera far from the filmed path (column~7).}
\label{fig:teaser}
\end{strip}

\begin{abstract}
Reconstructing an editable 3D scene from video requires matching visible structure while plausibly completing unseen surfaces. Agents can write scene-construction programs, but direct reconstruction attempts often produce errors in layout, scale, and lighting. We introduce \sys{}, an automated system for reconstructing mesh scenes from real and generated videos, enabling text-to-scene construction when paired with video generation. A software harness coordinates agents, pretrained vision models, and graphics tools through an explicit reconstruction protocol. Shared camera, depth, and segmentation estimates ground component construction: a scene-owner agent plans the environment and briefs authors, who write parameterized Blender programs. A visibility- and fit-based level-of-detail rule selects the construction level, while unseen regions are completed and marked as inferred. Assembled scenes use fitted lighting and supported motion, and revisions undergo programmatic checks and independent visual comparisons. Accepted programs and lessons are reused across scenes. We evaluate on 100 scenes; subset comparisons show improved original-view fidelity over direct-agent reconstruction and higher novel-view content similarity than the compared video-to-3D methods. We will release the harness, agent instructions, reusable libraries, and scenes.
\end{abstract}

\section{Introduction}
\label{sec:intro}

An editable 3D scene allows an environment to be explored, relit, and modified beyond a recorded or generated camera path. Recovering such a scene from video requires fitting visible structures and spatial relationships, separating materials from illumination, and completing unseen surfaces. Recent reconstruction and scene-generation models provide useful geometric estimates and priors~\citep{kerbl2023gaussians,shen2026lyra2,hunyuanworld2025,da3}. We build on these capabilities to reconstruct separate mesh objects, materials, and light sources from real and generated videos. Paired with text-to-video generation, the system also enables text-to-scene construction. Unseen surfaces are completed and marked as inferred.

Language and vision-language model agents can interpret structures, write Blender programs, and revise them from rendered feedback~\citep{hu2024scenecraft,sun2023threedgpt,yin2026viga,huang2024blenderalchemy}. Directly assigning the entire reconstruction task to an agent leaves it to establish cameras and scale, choose constructions, and assess its own result. In our zero-shot comparisons, two frontier agents produce recognizable content but simplified geometry and materials, with errors in layout, dimensions, and lighting (Fig.~\ref{fig:teaser}, columns 3 and 4). Some attempts project footage onto proxy surfaces, reproducing input views while exposing artifacts from new viewpoints. These failures motivate a process that grounds agent construction in geometric estimates and evaluates revisions in the assembled scene.

We introduce \sys{}, an automated system for coordinated component construction from video (Fig.~\ref{fig:method}). Pretrained vision models supply shared camera, depth, and segmentation estimates. A scene-owner agent plans components and briefs author agents, which return parameterized Blender generators. Agents choose scene-specific constructions; the method specifies their shared inputs, authoring interfaces, level-of-detail policy, and revision procedure. The level-of-detail policy uses visibility and fit to select procedural recipes, fitted instances, or part-by-part construction. Components are assembled with fitted lighting and supported motion, then revisions undergo programmatic checks and render comparisons by separate judge agents. A software \emph{harness} coordinates agents, models, and graphics tools and applies the acceptance rule. Accepted generators and curated lessons are reused in later runs. Section~\ref{sec:method} details this structure and the decisions delegated to agents.

Our contributions are:
\begin{itemize}
  \item An observation-grounded reconstruction framework that coordinates component authoring to produce editable scenes from real and generated videos, enabling text-to-scene construction when paired with video generation (Secs.~\ref{sec:overview}--\ref{sec:authoring}).
  \item A level-of-detail policy and scene-level revision procedure that allocate individual modeling according to visibility and fit, and evaluate changes through programmatic checks and independent visual comparisons (Secs.~\ref{sec:lod} and \ref{sec:loop}).
  \item An evaluation across 100 scenes, including direct-agent comparisons, a level-of-detail ablation, and open-weight author substitutions (Sec.~\ref{sec:exp}), accompanied by a planned release of code, instructions, libraries, scenes, and evaluation records.
\end{itemize}

\section{Related Work}
\label{sec:related}

\noindent\hspace*{\parindent}\textbf{Reconstruction from video.} Structure from motion and multi-view stereo recover cameras and surfaces~\citep{schonberger2016colmap,schonberger2016mvs,pan2024glomap}; feed-forward models predict point maps, depth, and poses~\citep{wang2024dust3r,wang2025vggt,wang2026pi3,da3,vipe}. Neural radiance fields~\citep{mildenhall2020nerf,barron2022mipnerf360} and Gaussian splatting~\citep{kerbl2023gaussians,jiang2025anysplat} represent appearance for view synthesis, with surface-oriented variants supporting mesh extraction~\citep{huang2024twodgs,chen2024pgsr}. Dynamic methods recover time-varying scenes~\citep{luiten2024dynamic,wu2024fourdgs,wang2025shapeofmotion}, while generative models synthesize explorable environments~\citep{shen2026lyra2,hunyuanworld2025,liu2026omniroam}. \sys{} uses a video-to-3D model as a measurement prior, then constructs separate, named mesh components with explicit materials, lighting, and supported animation.

\noindent\hspace*{\parindent}\textbf{Object-level reconstruction.} Methods retrieve and align CAD models to scans~\citep{avetisyan2019scan2cad}, or generate objects using learned priors from an image~\citep{huang2025midi,meng2025scenegen,ardelean2025gen3dsr,chen2025sam3d} or multiple views~\citep{ni2025dprecon,xia2025holoscene,xia2026simrecon,xia2026fire3d,schmid2026genrecon}. Video2Game reconstructs interactive mesh environments from video~\citep{xia2024video2game}. Our components are constructed by parameterized programs fitted to video observations, allowing their construction procedures and parameters to be revised and reused.

\noindent\hspace*{\parindent}\textbf{Agentic scene construction.} Language-model systems arrange assets~\citep{yang2024holodeck,feng2023layoutgpt}, write scene programs~\citep{sun2023threedgpt,hu2024scenecraft}, and coordinate generators with critics or verifiers~\citep{xia2026sage,erkoc2026worldagents,ling2025scenethesis}. Image-conditioned approaches predict structured scene descriptions~\citep{kulits2024igllm}, refine programs through rendering~\citep{yin2026viga}, orchestrate specialist agents~\citep{kim2026sceneconductor,sun2026harmony}, or edit existing scenes~\citep{huang2024blenderalchemy}. \sys{} grounds component authoring in shared video-derived measurements and evaluates revisions in a common assembled scene. Separate judges compare matched renders in both display orders, with identical-image controls and a prescribed acceptance rule. This procedure addresses the order sensitivity documented for language-model judges~\citep{zheng2023judging,wang2023fair}.

\noindent\hspace*{\parindent}\textbf{Procedural generation, reuse, and level of detail.} Infinigen generates natural and indoor environments procedurally~\citep{raistrick2023infinigen,raistrick2024infinigenindoors}. Voyager stores reusable agent programs~\citep{wang2024voyager}, and SceneCraft builds a scene-function library~\citep{hu2024scenecraft}. Our library retains generators fitted to video observations for adaptation to later scenes (Sec.~\ref{sec:libraries}). Rendering level-of-detail methods select representations according to viewing conditions~\citep{clark1976hierarchical,kerbl2024hierarchical}. Our policy instead allocates construction detail according to visibility and achieved fit; unresolved detail and unseen surfaces are completed procedurally and recorded as inferred (Sec.~\ref{sec:lod}).

\section{Method}
\label{sec:method}

Given a captured or generated video, \sys{} reconstructs an editable Blender scene containing separate mesh objects, materials, light sources, and animation data for supported moving objects.

\subsection{Overview and Terminology}
\label{sec:overview}

A \emph{reconstruction run} (or simply \emph{run}) is one complete attempt to reconstruct a scene; all agent sessions, vision model inferences and compute jobs for that attempt belong to the same run. %

\begin{figure*}[t]
\centering
\includegraphics[width=\textwidth]{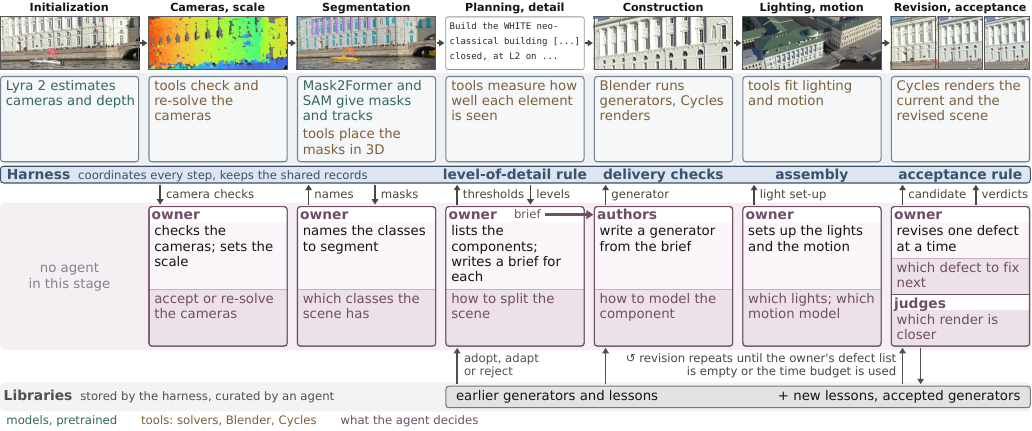}
\caption{The stages of \sys{}; the pictures are artefacts of the Winter Palace run. Above the band: each stage's \tableModel{models} and \tableTool{tools}. The band: the \tableHarness{harness} and its rules. Below: what each \tableAgent{agent} does and, in the darker cell, what it decides for the scene.}
\label{fig:method}
\end{figure*}

\noindent\hspace*{\parindent}\textbf{Scene representation and agent roles.}
The \emph{current scene} is the reconstruction state at a given stage of the run: its cameras and world coordinates, geometry, materials, lighting, and motion. A \emph{component} is an authored portion of the scene, such as a building or terrain surface. A \emph{generator} is a parameterized Blender program that constructs geometry and materials. A \emph{family} groups objects or parts that share construction logic. A family may span several objects, such as windows on different buildings. An \emph{instance} is one occurrence of an object or part in a family; different instances can have different generator parameters and poses.

We use three types of agents. A \emph{scene owner} agent coordinates the run, maintains the current scene, plans components, and assigns construction tasks to \emph{author} agents. Authors write and revise generators for particular families or spatial regions. Separate \emph{judge} agents compare candidate reconstructions with the current accepted scene and the observations. All three roles are language or vision-language model agents; our default owner uses Claude Opus~5.5, with alternative author models evaluated in Sec.~\ref{sec:openweight}.

 Figure~\ref{fig:method} shows the stages of a run and Algorithm~\ref{alg:run} summarizes it; the comment on each of its lines names who acts. Supp.~S8 lists the responsibilities by stage.

\begin{algorithm}[t]
\caption{One reconstruction run (notation in Supp.~S6). Agent actions are written in prose; small capitals group harness operations described in the text. The comment on a line names who acts, in the colours of Fig.~\ref{fig:method}.}
\label{alg:run}
\small
\begin{algorithmic}[1]
\REQUIRE selected shot $V$; libraries $L_g,L_e$; budget $B$
\ENSURE scene $S$ after the last accepted revision, or no scene
\STATE $O \leftarrow$ \textsc{InitializeObservations}$(V)$: frames, Lyra~2 cameras and depth \COMMENT{\tableHarness{harness}, \tableModel{models}} \label{ln:init}
\STATE owner validates cameras, aligns gravity, and estimates scale; updates $O$ \COMMENT{\tableAgent{owner}, \tableTool{tools}} \label{ln:cam}
\STATE segment and track with Mask2Former and SAM using owner prompts; locate masks in 3D; update $O$ \COMMENT{\tableModel{models}, \tableTool{tools}, \tableAgent{owner}} \label{ln:seg}
\STATE owner plans components; the level-of-detail rule proposes a level of detail per element \COMMENT{\tableAgent{owner}, \tableHarness{harness}} \label{ln:plan}
\STATE $S \leftarrow$ none; the first revision is the whole scene
\WHILE{time remains within budget $B$} \label{ln:loop}
  \STATE owner issues briefs for the components of the revision using $O,L_g,L_e$, or makes a small change itself \COMMENT{\tableAgent{owner}} \label{ln:brief}
  \STATE authors construct or revise generators $G$ until delivery checks pass \COMMENT{\tableAgent{authors}, \tableHarness{harness}} \label{ln:auth}
  \STATE $S' \leftarrow$ \textsc{Assemble}$(G,O)$ \COMMENT{\tableHarness{harness}} \label{ln:asm}
  \STATE reconsider detail with the level-of-detail rule; fit affected lighting, materials, and supported motion in $S'$ \COMMENT{\tableHarness{harness}, \tableTool{tools}, \tableAgent{owner}} \label{ln:fit}
  \IF{$S$ is none}
    \STATE \textbf{if} $S'$ passes the scene checks \textbf{then} $S \leftarrow S'$ \COMMENT{\tableHarness{harness}}
  \ELSIF{scene checks regress or motion check fails relative to $S$} \label{ln:check}
    \STATE reject $S'$ \COMMENT{\tableHarness{harness}}
  \ELSE
    \STATE render $S,S'$ with matched comparison settings \COMMENT{\tableTool{tools}} \label{ln:render}
    \STATE judges compare renders using reference frames in $O$ \COMMENT{\tableAgent{judges}}
    \STATE \textbf{if} \textsc{Accept}(verdicts, declared revision) \textbf{then} $S \leftarrow S'$ \COMMENT{\tableHarness{harness}} \label{ln:accept}
  \ENDIF
  \STATE owner compares renders of $S$ with frames in $O$ and lists defects; \textbf{if} the list is empty \textbf{then return} $S$ \COMMENT{\tableAgent{owner}}
  \STATE owner declares the next revision and its primary view group, and corrects derived observations in $O$ if needed \COMMENT{\tableAgent{owner}} \label{ln:next}
\ENDWHILE
\RETURN $S$
\end{algorithmic}
\end{algorithm}

\subsection{Video Processing and Geometric Initialization}
\label{sec:camera}
\noindent\hspace*{\parindent}\textbf{Frames and initial reconstruction.}
The \tableHarness{harness} extracts timestamped frames and detects shot boundaries; each run reconstructs one continuous shot. \tableModel{Lyra~2}'s reconstruction branch~\citep{shen2026lyra2} estimates cameras, depth, and a Gaussian representation, providing initial geometry for spatial measurements. These estimates serve as correctable priors rather than final scene geometry.

\noindent\hspace*{\parindent}\textbf{Camera validation and world coordinates.}
\tableTool{Geometric diagnostic tools} built with OpenCV and numerical routines, rather than large learned models, report reprojection error, trajectory discontinuities, vertical alignment, and near-object parallax (Supp.~S1). The \tableAgent{owner} uses these diagnostics to accept, re-estimate, or correct the cameras before construction. It establishes a gravity-aligned coordinate system and estimates scale from two independent recognizable size references, checking their agreement. All components use these shared cameras and coordinates.

\subsection{Segmentation and 3D Association}
\label{sec:association}
\noindent\hspace*{\parindent}\textbf{Segmentation and tracking.}
The \tableAgent{owner} names target classes and supplies text or box prompts. \tableModel{Mask2Former}~\citep{cheng2022mask2former} provides semantic regions, \tableModel{SAM~3.1}~\citep{carion2025sam3} segments prompted instances, and \tableModel{SAM~2.1}~\citep{ravi2024sam2} propagates instance masks through time. Each covers a case that the others handle poorly: \tableModel{Mask2Former} labels continuous far regions that prompted instances miss, and \tableModel{SAM~2.1} keeps instance identities in dense scenes (Supp.~S1). A \emph{track} associates masks of the same instance across frames. \tableAgent{Authors} obtain masks for constituent parts using the same prompting procedure.

\noindent\hspace*{\parindent}\textbf{Association with 3D geometry.}
\tableTool{Geometric tools} project reconstructed points into the views, filter occluded points, and accumulate labels from overlapping masks. Plane fitting estimates supporting surfaces and facades; clustering provides approximate instance locations and extents (Supp.~S1). This coarse 3D layout lets the owner plan the scene and assign authors, and lets assembly place each component at its estimated location instead of searching for its pose. These estimates inherit camera, depth, and segmentation errors and describe only observed geometry. For example, labeled building points constrain placement and visible dimensions; an author constructs coherent walls, a roof, and unseen sides.

\subsection{Scene Decomposition, Component Authoring}
\label{sec:authoring}
\noindent\hspace*{\parindent}\textbf{Scene planning and briefs.}
\label{sec:author}
The \tableAgent{owner} plans components, their relationships, and plausible completions of unseen parts, keeping continuous surfaces such as terrain within one component. Unseen sides and surroundings are built by the same generators as the observed parts and recorded as inferred (Supp.~S13). It builds a coarse environment, then delegates families or large structures to parallel authors. Delegation separates the work into bounded tasks: a family's author shares one generator and its lessons across all instances, and a smaller model can take such a task (Sec.~\ref{sec:openweight}; Supp.~S9). Each author receives a \emph{brief} specifying relevant observations, dimensions, coordinates, neighboring components, \emph{support constraints} (what the component rests on or attaches to), and proposed level of detail (Sec.~\ref{sec:lod}), together with current lighting, tools, and relevant generators and lessons.

\noindent\hspace*{\parindent}\textbf{Parameterized construction and fitting.}
\tableAgent{Authors} write Blender generators with parameters for dimensions, placement, orientation, and counts. They choose construction procedures and revise code and parameters by comparing renders with the footage, optionally invoking inverse-rendering tools. Components must remain separate and movable, include unseen sides, and use materials without copied footage pixels (Supp.~S5). Material parameters describe reflectance under the fitted illumination (Sec.~\ref{sec:lighting}).

\noindent\hspace*{\parindent}\textbf{Author delivery.}
Authors deliver generator code, named closed mesh objects, an object-label manifest, declared supports, and records of measurements and inferred parts. The \tableHarness{harness} checks deliveries and compares rendered silhouettes with observation masks. Failed deliveries return for revision; passing permits assembly. Acceptance of changes to the assembled scene follows Sec.~\ref{sec:loop}.

\subsection{Level of Detail}
\label{sec:lod}
The video supports different amounts of detail for different scene elements. Our level-of-detail policy chooses how each element is constructed from its own observations, and which detail is generated from regional statistics or structural assumptions. Components have a hierarchy of elements, such as a bed, its plants, and their leaves, or a district, its buildings, and their windows. A \emph{construction level} describes how an element is built; it is distinct from its depth in this hierarchy. Part-by-part construction can be applied recursively at successive tiers.

\noindent\hspace*{\parindent}\textbf{Construction levels.}
We use the following three levels, abbreviated L0, L1, and L2 in the author briefs:
\begin{itemize}
  \item \textbf{L0: Procedural recipe.} Generate the element on its support using density, size, and color estimated for its region.
  \item \textbf{L1: Fitted instance.} Fit a generator instance's position, scale, height, and orientation to the element's own masks.
  \item \textbf{L2: Part by part.} Construct observed parts using their individual masks and spatial measurements, applying the policy recursively to those parts.
\end{itemize}
For example, a plant can be generated from regional statistics, fitted to its own outline, or built from individually observed leaves. The level is proposed during scene planning and reconsidered at each revision. Two visibility tests propose the construction level before building; a subsequent fit test checks a constructed part-by-part candidate.

\noindent\hspace*{\parindent}\textbf{Visibility-based proposal.}
Let $\mathcal{T}(e)$ be the frames in which element $e$ has an instance mask, and $M_t(e)$ its mask in frame $t$. Its frame share is $a_t(e)=|M_t(e)|/(WH)$ for a $W{\times}H$ image, and $\tilde a(e)$ is the median share over $\mathcal{T}(e)$. Let $m(e)$ count the masks of its parts over all frames of $\mathcal{T}(e)$, and let $w(e)$ be the median diameter of the largest circle inscribed in each of these masks. Parts are not tracked at this stage, so a part visible in $k$ frames contributes $k$ masks.

\tableTool{Tools} compute these quantities. Before construction, the \tableAgent{owner} sets and records the visibility thresholds for each family, with a reason; defaults are in Supp.~S3. The instance-visibility test $S_1$ requires enough views and sufficient median frame share. The part-visibility test $S_2$ additionally considers sustained image size, part width, and the number of part masks:
\begin{align}
S_1(e):\;& |\mathcal{T}(e)|\ge n_1,\;\; \tilde a(e)\ge\alpha_1, \label{eq:s1}\\
S_2(e):\;& \tilde a(e)\ge\alpha_2,\;\; |\{t\in\mathcal{T}(e): a_t(e)\ge\alpha_0\}|\ge n_2, \nonumber\\
 & w(e)\ge w_0,\;\; m(e)\ge m_0. \label{eq:s2}
\end{align}
An element that passes both tests is proposed as a part-by-part candidate, one that passes only $S_1$ as a fitted instance, and any other as a procedural recipe. These tests use image measurements, not agent visual verdicts; their outputs depend on the supplied masks and recorded thresholds.

\noindent\hspace*{\parindent}\textbf{Construction, fitting, and fallback.}
For a proposed L2 construction, the \tableAgent{author} obtains part masks with SAM~3.1 and constructs and refines a candidate $G_e$. Authors deliver a generator shared by the family and a parameter record for each element; fitting tools adjust parameters against the corresponding masks. At L1 this fits the whole instance, optionally using consistent depth for bounded deformation; at L2 the individual parts have their own masks and measurements.

Let $R_t(G)$ be the silhouette of a construction $G$ in a render of the assembled scene at frame $t$, in which occluders hide the same pixels as in the observed mask $M_t(e)$, and define $\mathrm{IoU}_t(e,G)=|R_t(G)\cap M_t(e)|/|R_t(G)\cup M_t(e)|$. The fit test requires median silhouette intersection-over-union of at least $\tau=0.6$ over the element's track:
\begin{equation}
F(e):\;\operatorname*{med}_{t\in\mathcal{T}(e)}\mathrm{IoU}_t(e,G_e)\ge\tau.
\label{eq:fit}
\end{equation}
This is a silhouette-consistency check; agreement with the whole element's outline alone does not establish the accuracy of its internal parts.

\tableTool{Tools} combine visibility and achieved fit into the level that the policy proposes for an element:
\begin{equation}
\ell(e)=\begin{cases}\text{part by part} & S_1(e)\wedge S_2(e)\wedge F(e),\\ \text{fitted instance} & \text{otherwise, if } S_1(e),\\ \text{procedural recipe} & \text{otherwise.}\end{cases}
\label{eq:level}
\end{equation}
A part-by-part candidate that fails $F$ is thus proposed as a fitted instance. The same policy is applied recursively to retained parts, so instances of one family can have different construction levels. The \tableAgent{author} decides what is built: the instructions allow it to build individual parts wherever those parts have their own masks, also inside a fitted instance. The record for each element therefore holds the measurements, the proposed level, and the built level.

The instructions emphasize separability as well as image size: a small, cleanly outlined dormer can be fitted individually, whereas unresolved house fronts are represented by their block's estimated height, roof line, and window spacing. The visibility tests do not measure separability: the author judges it on the masks while building, and builds an element whose outline merges with its neighbors at a lower level than proposed. Detected elements remain represented even if their masks or tracks cannot support individual fitting.

\noindent\hspace*{\parindent}\textbf{Completion and novel-view detail.}
Below the built level, procedural rules generate unresolved detail using estimated density, size, and color. Structural assumptions and the scene plan provide plausible completions of unseen parts. Generated detail and unseen completions are recorded as inferred. If an element's placement is uncertain, it remains generated within its parent's constraints rather than being dropped.

After assembly, a tool compares each element's maximum projected size over the delivered views with its size in the input frames. Where a novel view comes closer than any input view, the author generates one additional tier of finer elements. This tier is also inferred: a closer rendered view does not supply new observations of the scene.

\begin{figure*}[t]
\centering
\includegraphics[width=\textwidth]{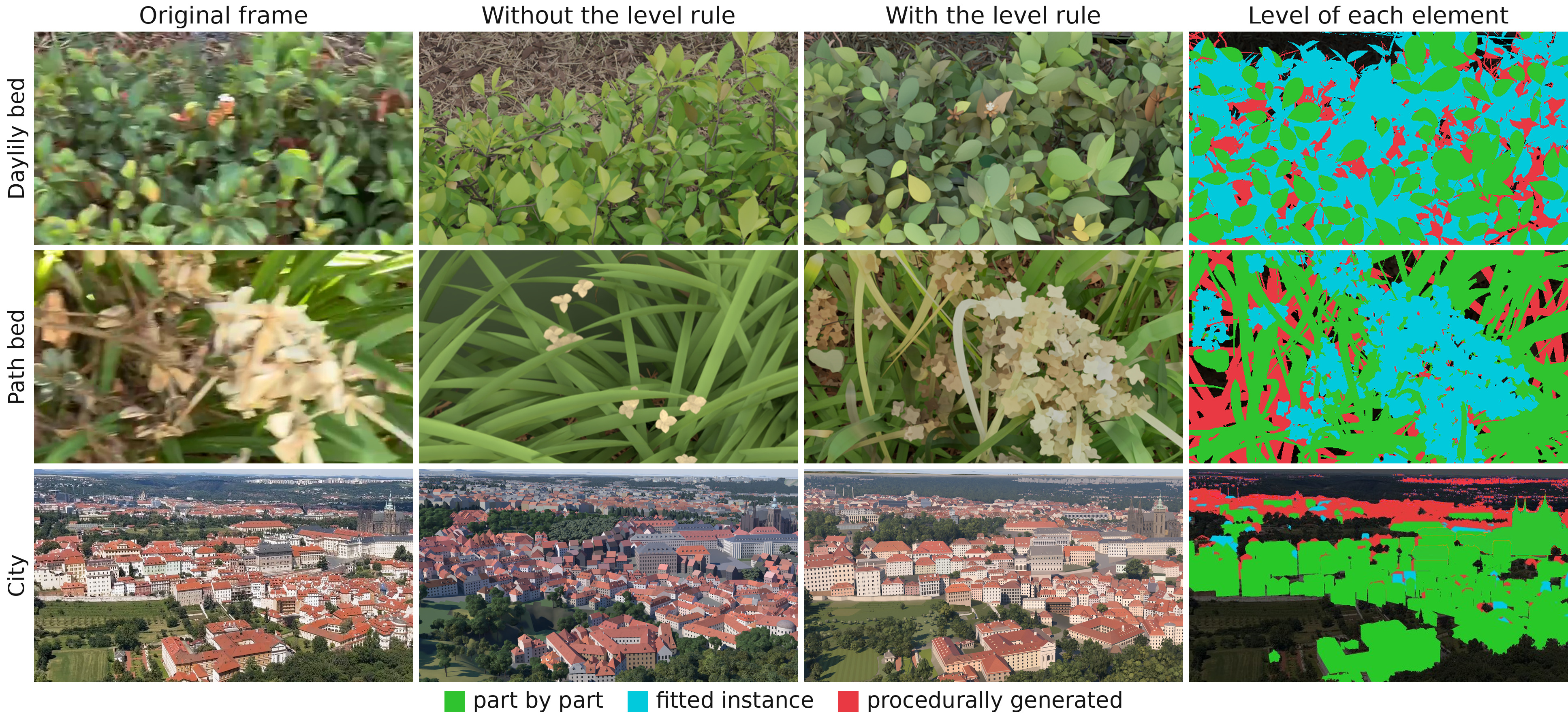}
\caption{The level-of-detail rule (Eq.~\ref{eq:level}) at two scales: two garden beds and a city filmed from a hill (last row). Without the rule, every plant of a bed is one instance of its family's generator: it has leaves of the right kind but not the leaves of the frame. In the city frame, the elements built part by part cover 77\% of the building pixels.}
\label{fig:lod}
\end{figure*}

Figure~\ref{fig:lod} illustrates the policy at plant and city scales. In the city, whose original frame contains about 1,800 buildings, near landmarks are built individually and part by part, further houses are fitted to roof outlines, and unresolved blocks are generated inside their observed regions. On the embankment, theatre columns are built individually, the distant palace row uses fitted instances, and the city behind the frontage uses inferred recipes.

\subsection{Scene Assembly, Fitting, and Revision}
\label{sec:loop}
The owner assembles components in the shared coordinate system and fits lighting, materials, and supported motion. The initial assembly must pass the scene checks below. Thereafter, the \emph{parent} is the current accepted scene and a \emph{candidate} is a proposed revision, evaluated in the assembled environment to account for changes to shadows, occlusions, and neighboring components.

\noindent\hspace*{\parindent}\textbf{Lighting and materials.}
\label{sec:lighting}
The \tableAgent{owner} selects light sources and calibration regions for fitting their strengths. Outdoor scenes use sun and sky; indoor scenes use visible luminaires, with windows and doorways as sky portals. Fitting \tableTool{tools} sweep sun directions against observed lit-and-shaded regions, initialized from date, time, and location when available. Authors choose material models; diffuse albedo is estimated under the fitted illumination and checked for consistency between lit and shaded pixels. Inconsistency triggers lighting refitting. Reflective and transmissive surfaces use corresponding material models.

A fitted camera model maps scene-linear renders to displayed frames through exposure, white balance, and a bounded tone curve (Supp.~S1). Its parameters are held fixed while lighting or material revisions are compared, preventing candidate-specific exposure adjustments from compensating for reconstruction errors.

\noindent\hspace*{\parindent}\textbf{Motion fitting.}
\label{sec:motionmethod}
The \tableAgent{owner} selects a rigid-body, mechanism, or water-surface representation; fitting \tableTool{tools} estimate its parameters. Rigid poses are fitted to tracked silhouettes through the shared cameras, with trajectories constrained by supports. Mechanisms use axes, periods, and speeds estimated from the sequence; water uses a wave surface fitted to its extent, speed, and pattern. Animation uses the original timestamps, and shape changes trigger pose refitting.

\noindent\hspace*{\parindent}\textbf{Candidate revisions and checks.}
The \tableAgent{owner} compares the current scene with the original frames, lists defects, and proposes or delegates a revision targeting the most severe defect. Each comparison evaluates one declared revision together with its necessary refits. Missing or inconsistent observations can trigger additional segmentation, measurement, or correction of derived estimates; the original frames remain the reference for image agreement.

Before comparison, the level-of-detail policy is reconsidered for affected elements and lighting and supported motion are refitted as needed. The \tableHarness{harness} runs support and physical audits covering attachment relationships, interpenetration, emission from non-light materials, painted reflections, and implausible albedo; all but painted reflections, which the owner inspects, are tested by programs. Each audit counts findings on the assembled scene, and a candidate may not have more findings of a kind than its parent (Supp.~S8). For every moving object, median tracked-mask IoU may decrease by at most 0.005. A candidate that fails these scene or motion checks is rejected before visual judging.

\noindent\hspace*{\parindent}\textbf{Visual comparison and acceptance.}
Before judging, the \tableAgent{owner} declares the render pairs and the primary \emph{view group}, a set of pairs for one kind of view, such as original or novel views. Candidate and parent use a common camera table, timestamps, observations, and rendering settings. Separate \tableAgent{judges}, without construction history or candidate identities, see a reference frame and two renders and choose the closer reconstruction or a tie. Judges are separate because agents accept their own work too readily, and the comparison controls two measured judge biases: judges pick a side even between identical images, and prefer the richer background (Supp.~S2). Novel views use available reference frames to assess consistency and plausibility, not the correctness of unseen geometry.

Each pair is evaluated in both display orders and counts as a win only when the same reconstruction wins both; a pair with a tie or with disagreeing orders is left out of the test. Identical-image controls test preferences unrelated to scene differences. The \tableHarness{harness} accepts a candidate when it wins the primary group under a two-sided sign test over pairs at significance level 0.05, controls tie, and no other evaluated group is lost, that is, won by the parent under the same test. Supp.~S2 describes the judging and early-stopping procedure.

Corrections that the owner declares, before judging, as required for physical validity undergo the same audits and motion checks but are retained unless the parent wins a view group under the same test. Accepted candidates replace the parent; rejected candidates leave it unchanged. A run returns its latest accepted scene when the owner has no further listed defects or the time budget expires; if no initial assembly passes the checks, it returns no scene (Algorithm~\ref{alg:run}).

\noindent\hspace*{\parindent}\textbf{Reuse across runs.}
\label{sec:libraries}
The generator library $L_g$ stores accepted construction programs by family. A curation agent distills construction lessons into the experience library $L_e$, recording applicability conditions and supporting observations. Later owners and authors retrieve relevant generators and lessons and adapt them to the new video's observations. Reused generators undergo the same fitting, checks, and acceptance process as new ones; reuse does not update model weights. Generators are written for a family rather than for one view, which also lets a scene be extended beyond the filmed region.

\section{Results}
\label{sec:exp}

\subsection{Setup}
\label{sec:setup}
\noindent\hspace*{\parindent}\textbf{Scenes and evaluation subsets.}
\label{sec:dataset}
We reconstruct 100 scenes: 39 static and 20 moving-element Wikimedia Commons videos, 17 generated videos, and 24 photo sets and our own captures, including Mip-NeRF~360~\citep{barron2022mipnerf360}, LLFF~\citep{mildenhall2019llff}. The videos span indoor and outdoor scenes captured by handheld camera and by drone. Different experiments use subsets of this collection. Video-to-3D methods were run on 21 scenes (8 real, 13 generated). \sys{} returns a scene on all 21; the reconstruction of a compared method can be registered to the original cameras on 19 scenes for Pi3-X, 15 for PGSR, and 12 for COLMAP (Supp.~S12). Table~\ref{tab:metrics} reports the 10 jointly aligned scenes (4 real, 6 generated). Direct-agent comparisons cover 30 scenes.

\noindent\hspace*{\parindent}\textbf{Metrics.}
For video-to-3D comparisons, we report LPIPS~\citep{zhang2018lpips} and DINOv3 feature similarity~\citep{simeoni2025dinov3} against the input video at up to 8 original cameras per scene; these frames are among the input of every method, and the table averages per-scene means. Direct-agent comparisons use 101 frames across 30 scenes. At 8 novel cameras per scene, spaced evenly on a path around the scene (Supp.~S12), DINOv3 similarity to the best-matching original frame measures content similarity, not geometric accuracy. Pairwise comparisons follow Sec.~\ref{sec:loop}, using Claude Opus~5.5 judges; a pairwise win requires agreement across both display orders.

\noindent\hspace*{\parindent}\textbf{Baselines.}
Direct-agent baselines are GPT-6 Astra and Claude Opus~5.5 at maximum reasoning effort, given access to the video and Blender without our harness, supplied measurements, fitting tools, or acceptance procedure. Each agent iterates on its scene until it considers the result complete. Video-based baselines use their released pipelines: Pi3-X~\citep{wang2026pi3}, PGSR~\citep{chen2024pgsr}, COLMAP dense reconstruction and meshing~\citep{schonberger2016colmap,schonberger2016mvs}, and Video2Game~\citep{xia2024video2game} (Fig.~\ref{fig:orbit} only). Gaussian reconstructions, including our Lyra~2 prior, are shown in Fig.~\ref{fig:teaser} and Supp.~S9. Single-image generators receive one video frame: MIDI-3D~\citep{huang2025midi}, SceneGen~\citep{meng2025scenegen}, Gen3DSR~\citep{ardelean2025gen3dsr}, and SAM~3D Objects~\citep{chen2025sam3d} (Supp.~S10).

\noindent\hspace*{\parindent}\textbf{Output-token cost.}
On 40 Wikimedia Commons scenes (20 static and 20 with moving elements), owners, authors, and judges produce 0.9--5.6 million output tokens per run (median 2.4 million). At US \$~20 per million output tokens, this corresponds to \$~18--111 per scene (median \$~47). These figures exclude compute for vision models, fitting, and rendering (which can be run locally on the user's compute hardware).

\subsection{Comparison with Other Methods}
\label{sec:prior}
\noindent\hspace*{\parindent}\textbf{Direct-agent reconstruction.}
On 30 scenes and 101 original-camera frames, Claude Opus~5.5 and GPT-6 Astra obtain LPIPS 0.66 and 0.69 and DINOv3 similarity 0.55 and 0.52, compared with 0.51 and 0.76 for \sys{} (lower LPIPS and higher DINOv3 similarity is better). Figure~\ref{fig:teaser} illustrates errors in their layout, scale, lighting, and geometric detail. These results demonstrate the benefit of our complete reconstruction protocol and its supplied models and tools.

\begin{figure*}[t]
\centering
\includegraphics[width=\textwidth]{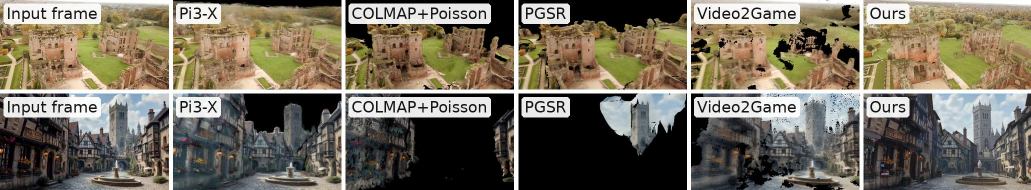}
\caption{Comparison with released video-based reconstruction methods on a real drone video (top: Kenilworth Castle) and a generated video (bottom: an old street). Every tile is at the camera of the original frame on the left; ours is on the right. Black: no output.}
\label{fig:orbit}
\end{figure*}

\noindent\hspace*{\parindent}\textbf{Video-to-3D reconstruction.}
On the 10 jointly aligned scenes, \sys{} has lower original-view LPIPS (0.50) than the compared methods (best baseline 0.56) and ties Pi3-X in DINOv3 similarity (0.74). At novel cameras, its content similarity to the input video is higher (0.53 versus at most 0.38; Table~\ref{tab:metrics}). Figure~\ref{fig:orbit} compares video-based methods at an original camera; single-image results are in Supp.~S10.

\begin{table}[t]
\caption{Geometry-producing video-to-3D methods at the original cameras (up to 8 frames per scene) and at 8 novel cameras (DINOv3 against the best-matching original frame), on the 10 scenes (4 real, 6 generated) on which every method's output could be aligned with the original cameras.}
\label{tab:metrics}
\footnotesize
\centering
\setlength{\tabcolsep}{5pt}
\begin{tabular}{@{}lccc@{}}
\toprule
 & \multicolumn{2}{c}{original views} & novel views \\
\cmidrule(lr){2-3}\cmidrule(lr){4-4}
Method & LPIPS $\downarrow$ & DINOv3 $\uparrow$ & DINOv3 $\uparrow$ \\
\midrule
Pi3-X & 0.56 & \textbf{0.74} & 0.38 \\
PGSR & 0.61 & 0.54 & 0.31 \\
COLMAP dense + mesher & 0.67 & 0.41 & 0.27 \\
Ours & \textbf{0.50} & \textbf{0.74} & \textbf{0.53} \\
\bottomrule
\end{tabular}
\end{table}
\begin{figure*}[t]
\centering
\includegraphics[width=\textwidth]{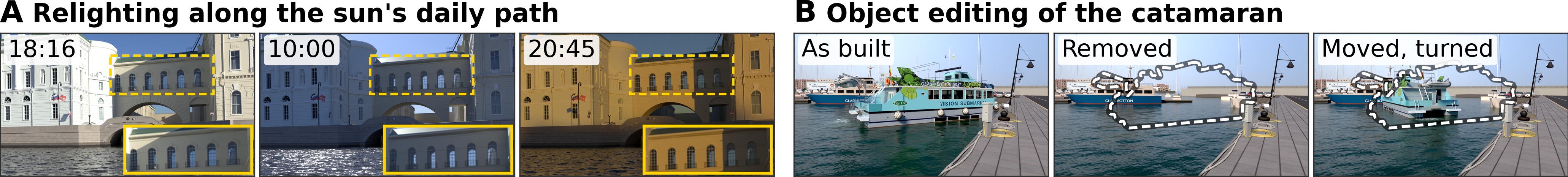}
\caption{Applications; all images are renders of our scenes. (A)~Winter Palace embankment: the sun fitted to the footage's shadows (18:16), then moved along that day's path to 10:00 and 20:45. (B)~Corralejo harbour: the catamaran as built, removed, and moved; the scene behind it continues.}
\label{fig:applications}
\end{figure*}

\subsection{Level-of-Detail Study}
\label{sec:lessons}
\noindent\hspace*{\parindent}\textbf{Level of detail.}
\label{sec:garden}
Three garden beds, each filmed in one hand-held shot, test the policy at leaf scale; the densest contains about 150 weeds, shrubs, and ferns. The policy yields lower plant-region LPIPS than one generator instance per plant on all three beds and on every evaluated frame (Table~\ref{tab:design}; Fig.~\ref{fig:lod}). For the city, building-pixel LPIPS drops from 0.53 to 0.49. Rendered plants cover 0.78 of the reference foliage pixels on the fence bed and 0.84 on the path bed. On the fence bed, revisions adding at least 300 observed leaves won their blind comparisons (8:0, 8:0, and 10:0).

\begin{table}[t]
\caption{Ablation of the level-of-detail rule: LPIPS $\downarrow$ on the pixels of the elements the rule governs.}
\label{tab:design}
\footnotesize
\centering
\setlength{\tabcolsep}{4pt}
\begin{tabular}{@{}lcccc@{}}
\toprule
 & fence bed & daylily bed & path bed & city \\
\midrule
one generator per element & 0.64 & 0.65 & 0.70 & 0.53 \\
level-of-detail rule (ours) & \textbf{0.55} & \textbf{0.57} & \textbf{0.58} & \textbf{0.49} \\
\bottomrule
\end{tabular}
\end{table}

\subsection{Author-Model Alternatives}
\label{sec:openweight}
We test DeepSeek V4.1 Flash and Nemotron 3.5 Super VL authoring for the same 14 scenes, with the owner model, delivery requirements, and judges unchanged. While the overall quality is lower (on the 11 scenes that also have a run with the Claude Opus~5.5 author, mean LPIPS and DINOv3 similarity at the original cameras are 0.55 and 0.66 for DeepSeek and 0.58 and 0.60 for Nemotron, against 0.47 and 0.79), the resulting scenes retain separate mesh objects, fitted lighting, and qualitative coherence at novel views (Supp.~S9). All 28 runs delivered a scene.

\subsection{Motion and Editing}
\label{sec:motion}
Moving objects are animated: the owner identifies the kind of motion and tools fit its parameters to the tracked masks. For example, a high-speed train, a harbor boat, a funicular car, an airliner, and a slewing crane follow the footage at median tracked-mask IoU 0.89--0.98 and median centroid error 2--13~px, measured over the frames in which each is tracked. Explicit animation allows the fitted motion to replay from novel cameras (Supp.~S11).

\label{sec:applications}
Separate meshes and explicit light sources also permit relighting (Fig.~\ref{fig:applications}A) and removal and repositioning of objects (Fig.~\ref{fig:applications}B). Surfaces revealed by an edit remain present, although their unseen geometry is inferred.

\subsection{Limitations}
\label{sec:limitations}
Best results require commercial models; current open-source author agents produce lower quality scenes. Reconstruction of unseen regions is plausible but generally shows lower quality and detail than observed regions. Our method is built on top of imperfect camera, segmentation, depth, and material estimates. Each run reconstructs one continuous shot; unifying shots across cuts remains future work. Human reconstruction is a hard researcch problem and we do not attempt it; non-rigid motion other than water surfaces remains outside the current scope.

\section{Conclusion}
\looseness=-1 We presented \sys{}, an automated system in which a software harness coordinates agents, pretrained vision models, and graphics tools to turn real and generated videos into editable 3D scenes. The agents decide what to build and which revisions to try; the harness keeps a revision if it passes programmatic checks and separate judge agents prefer its renders. At novel views, our scenes agree better in content with the footage than those of three video-to-3D methods that produce geometry, and they remain coherent when smaller open-weight models author the components.

\section*{LLM Usage Statement}
Claude (Fable~5.2 and Opus~5.5) and OpenAI Codex (GPT-6 Astra and GPT-6.1 Sol) were used in code development and experiment execution, and to assist with manuscript editing. Language-model agents and generated input videos are part of the method and its evaluation (Secs.~\ref{sec:method} and~\ref{sec:setup}). The authors take full responsibility for the paper's claims and results.

\onecolumn
\supppage{1}
\supppage{2}
\supppage{3}
\supppage{4}
\supppage{5}
\supppage{6}
\supppage{7}
\supppage{8}
\supppage{9}
\supppage{10}
\end{document}